\documentclass[review]{elsarticle}

\usepackage{lineno}
\usepackage{times}
\usepackage{epsfig}
\usepackage{graphicx}
\usepackage{amsmath}
\usepackage{amssymb}
\modulolinenumbers[5]
\DeclareMathOperator*{\argmax}{argmax}

\journal{Neural Networks}
\usepackage{multirow}
\usepackage{booktabs}
\usepackage{hhline}
\usepackage{dashrule}
\usepackage{array}
\usepackage{color, colortbl}
\definecolor{Gray}{gray}{0.9}
\usepackage{algorithm,algpseudocode}
\usepackage{lipsum}

\begin{document}

\begin{frontmatter}

\title{Self-Augmentation: Generalizing Deep Networks to Unseen Classes for Few-Shot Learning\tnoteref{mytitlenote}}

\author[address_1]{Jin-Woo Seo\fnref{equal}}
\author[address_1]{Hong-Gyu Jung\fnref{equal}}
\author[address_2]{Seong-Whan Lee\corref{mycorrespondingauthor}}
\fntext[equal]{Equal contribution}
\cortext[mycorrespondingauthor]{Corresponding author}
\ead{sw.lee@korea.ac.kr}

\address[address_1]{Department of Brain and Cognitive Engineering, Korea University, Anam-dong, Seongbuk-gu, \\ Seoul, 02841, Korea}
\address[address_2]{Department of Artificial Intelligence, Korea University, Anam-dong, Seongbuk-gu, \\ Seoul, 02841, Korea}

\begin{abstract}
Few-shot learning aims to classify unseen classes with a few training examples. While recent works have shown that standard mini-batch training with carefully designed training strategies can improve generalization ability for unseen classes, well-known problems in deep networks such as memorizing training statistics have been less explored for few-shot learning. To tackle this issue, we propose self-augmentation that consolidates self-mix and self-distillation. Specifically, we propose a regional dropout technique called self-mix, in which a patch of an image is substituted into other values in the same image. With this dropout effect, we show that the generalization ability of deep networks can be improved as it prevents us from learning specific structures of a dataset. Then, we employ a backbone network that has auxiliary branches with its own classifier to enforce knowledge sharing. This sharing of knowledge forces each branch to learn diverse optimal points during training. Additionally, we present a local representation learner to further exploit a few training examples of unseen classes by generating fake queries and novel weights. Experimental results show that the proposed method outperforms the state-of-the-art methods for prevalent few-shot benchmarks and improves the generalization ability.
\end{abstract}


\end{frontmatter}


\section{Introduction}
Deep networks have achieved remarkable performance in recognition problems \cite{krizhevsky2012imagenet,simonyan2014very,szegedy2015going,he2016deep, zhu2020learning, cheng2020face} over hand-crafted features \cite{lowe2004distinctive,roh2007accurate,bay2008speeded,dalal2005histograms,roh2010view,kang2014nighttime}. Assuming a large-scale training dataset is available, most studies focus on training deep networks on base classes to test unseen \textit{images} of trained classes. However, there is a growing interest in mimicking human abilities such as generalizing a recognition system to classify \textit{classes} that have never been seen before \cite{vinyals2016matching, liu2020label}. In particular, few-shot learning assumes only a few training examples are available for unseen classes. This is a challenging problem since it is highly possible that a few training examples will lead to network overfitting.

One paradigm for this challenge is meta-learning \cite{vinyals2016matching,snell2017prototypical,finn2017model}, where a large-scale training set for base classes is divided into several subsets (typically called tasks) and the network learns how to adapt to those tasks. In each task, only a few training examples are given for each class to mimic the environment of a test set for unseen classes.

Meanwhile, recent works have shown that a network trained with standard supervision can produce reasonable performance on unseen classes \cite{lifchitz2019dense,gidaris2019boosting,dvornik2019diversity}. In the training phase, this paradigm trains a network using a mini-batch sampled from a large-scale training dataset. In the test phase, unseen classes with a few training examples are evaluated using the same network. Thus, the goal is to develop a network that is generalizable to unseen classes by fully utilizing the knowledge learned from base classes.

Both paradigms share commonalities in that they leverage a large annotated collection. However, the following notable difference exists: Meta-learning learns to adapt quickly to new tasks by splitting base classes into several different tasks, whereas the standard supervision constructs a parameter space in which the \textit{unseen classes} can be identified using only the information for classifying the base classes at once. While the latter paradigm is closely related to classifying the \textit{unseen images} belonging to the base classes, only a few studies have taken advantage of lessons learned from the classical classification problem \cite{lifchitz2019dense,gidaris2019boosting,dvornik2019diversity}.

To tackle this issue, we take a closer look at the generalization ability of deep networks for few-shot learning. It is known that deep networks tend to have almost zero-entropy distributions as the softmax output produces one peaked value for a class \cite{verma2018manifold}. This overconfidence can occur even with randomly labeled training data as deep networks are likely to just memorize the training statistics \cite{thulasidasan2019mixup}. In our problem setting, this memorization property directly affects the performance on unseen classes as we rely heavily on the network ability trained on the dataset of base classes. The problem even worsens as we cannot apply a simple transfer learning strategy given that we have only a few training examples for unseen classes. Thus, to overcome the memorization issues, it is important to induce uncertainty in predictions about input images and regularize the posterior probability \cite{pereyra2017regularizing,devries2017improved,yun2019cutmix,zhang2018deep}.

With this in mind, we propose self-augmentation that incorporates regional dropout and knowledge distillation to improve the generalization ability\footnote{Henceforth, we denote the term ``generalization'' as the ability to adapt to unseen classes, given a network trained on base classes.} for few-shot leaning. Here, we use the self-augmentation term as we use input and output resources of the network itself to augment the generalization ability. Specifically, as one of the data augmentation techniques, we employ regional dropout, which substitutes a patch of an input image into other values such as zeros \cite{devries2017improved}, a patch of another image \cite{yun2019cutmix}, and another patch of the input image. We call the last regional dropout ``self-mix'' as it exchanges different patches of the input image itself. With this dropout effect, the generalization ability is improved as it prevents us from learning specific structures of a dataset. Furthermore, we found that an explicit regularization for the posterior probability is necessary to search for a proper manifold for unseen classes. To be specific, we utilize a backbone network that has auxiliary branches with its own classifier to enforce knowledge sharing. This sharing of knowledge forces each branch not to be over-confident in its predictions, thus improving the generalization ability. Cooperating with regional dropout, the experimental results show that knowledge distillation significantly boosts the performance on unseen classes.

Lastly, we present a fine-tuning method to exploit a few training examples given for unseen classes. As we train a network on base classes, we have the opportunity to improve the discriminative ability of the network for unseen classes using only 1 or 5 training examples.

To sum up, our main contributions are as follows: 
\begin{enumerate}
\item[1)] We present self-augmentation as a training framework to improve the generalization ability of deep networks. Specifically, we design consolidating regional dropout and knowledge distillation, which are less explored in the few-shot learning area.
\item[2)] We show that the newly proposed regional dropout, called self-mix, produces state-of-the art results when cooperating with knowledge distillation.
\item[3)] Lastly, we present a novel fine-tuning method, called a local representation learner, to exploit a few training examples of unseen classes, and show that the method improves the performance for all the few-shot learning benchmarks.
\end{enumerate}

\section{Related Work}
\subsection{Few-Shot Learning}
The literature on few-shot learning considers training and test datasets that are disjoint in terms of classes. Depending on how the training set is handled, we can categorize it into two main branches: meta-learning and standard supervised learning.

Meta-learning approaches train a network by explicitly emulating the test environment for few-shot learning. Using a training dataset, $n$ classes are randomly chosen with $k$ training examples, and $T$ queries are also randomly picked. Then, learnable parameters are obtained from the $n \cdot k$ training examples, and a loss is generated using the $T$ queries. A network is learned to reduce the loss by repeating this task several times. As a result, meta-learning  warms a network up to classify unseen classes with a few examples. Three approaches exist for this paradigm: 1) Metric-learning to reduce the distance among features of different classes \cite{snell2017prototypical,vinyals2016matching,sung2018learning,oreshkin2018tadam,li2019finding}, 2) optimization-based approaches to initialize a parameter space so that a few training examples of unseen classes can be quickly trained with the cross-entropy loss \cite{finn2017model,rusu2018meta,sun2019meta}, and 3) weight generation methods to directly generate classification weights used for unseen classes \cite{gidaris2018dynamic,gidaris2019generating,qi2018low}.

In contrast, the standard supervised learning approaches train a network as usual without splitting a training dataset into several tasks. In other words, this approach utilizes the training dataset as in the classical classification problem, but it aims to generalize unseen classes. To achieve this, dense classification applies the classification loss to all spatial information of an activation map to maximally exploit local information \cite{lifchitz2019dense}. A previous study used self-supervision and showed that the auxiliary loss without labels can extract discriminative features for few-shot learning \cite{gidaris2019boosting}. An ensemble method using multiple networks was also proposed to resolve the high-variance issue in few-shot learning \cite{dvornik2019diversity}.
\subsection{Generalization}
Many efforts have been made to understand the generalization performance of deep learning \cite{zhang2016understanding,guo2017calibration,pereyra2017regularizing,chaudhari2016entropy,verma2018manifold,neyshabur2017exploring,thulasidasan2019mixup,zhang2018deep, ukita2020causal}. Notably, it has been shown that deep networks easily adapt to random labels and are even well trained for images that appear as nonsense to humans \cite{zhang2016understanding}. Along the same lines, many works have found that deep networks produce overconfident classification predictions about an input, thus causing loss in the generalization performance \cite{pereyra2017regularizing,zhang2018deep,muller2019does,thulasidasan2019mixup}. To resolve this issue, recently, regional dropout \cite{devries2017improved,yun2019cutmix} and mixing up of two images \cite{zhang2017mixup,tokozume2018between} have been proposed as data augmentation techniques. Other researchers showed that label smoothing \cite{szegedy2015going} and knowledge distillation \cite{hinton2015distilling,zhang2018deep,zhu2018knowledge,sun2019deeply} effectively mitigate the overfitting problem by regularizing the posterior probability. In this paper, we expand these findings and indicate that perturbing input and output information should be extensively investigated for few-shot leaning. To this end, we propose a training framework that consolidates regional dropout and knowledge distillation, and further present a novel regional dropout called self-mix. In addition, we show that a novel fine-tuning method can be used to boost the performance of few-shot learning.

\begin{figure*}[t]
\begin{center}
\includegraphics[width=1.0\linewidth, height=43.5mm]{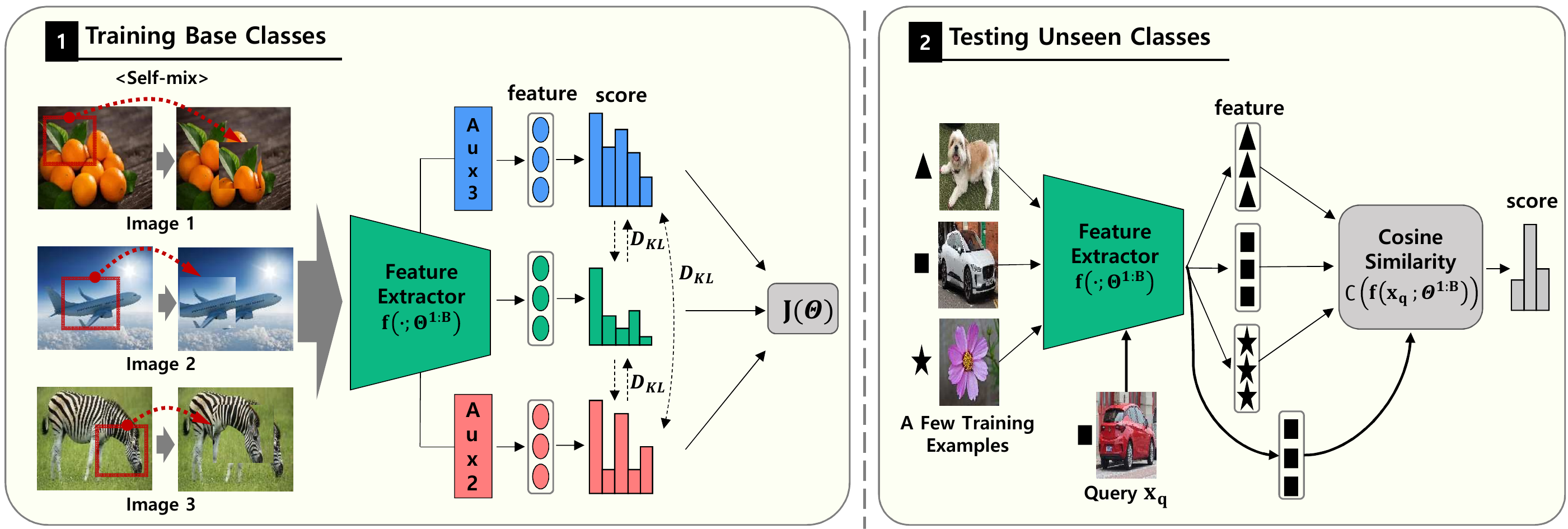}
\end{center}
   \caption{\textbf{Overview of the proposed self-augmentation framework.} The main network consists of three classifiers, two of which are derived from intermediate layers of the main branch. In the training phase, we first apply regional dropout to input images, which removes a part of the image by replacing it with other values. All the classifiers try to learn a more generalizable parameter space by minimizing the cross entropy loss and regularizing their prediction scores to have a similar distribution via the KL divergence. For inference, we simply use the main classifier to evaluate images from unseen classes. The right figure shows the case of $3$-way $1$-shot learning as an example.}
\label{fig:Overall_arc}
\end{figure*}

\section{Methodology}
\subsection{General Framework}
In this paper, we are interested in training a network on base classes to be generalizable to unseen classes. Before elaborating on the proposed method, we introduce the general framework for training and inference.

\subsubsection{Training}
We define a classifier as $C\left(f\left(\cdot ; \Theta^{1:B}\right)\right)$, where $f(\cdot ; \Theta^{1:B})$ is a feature extractor. Here, we denote the parameters from Block $1$ to $B$ as $\Theta^{1:B}$, assuming that we use a block-wise network such as ResNet \cite{he2016deep}. For the classifier, we use the cosine similarity that has been exploited for few-shot learning \cite{gidaris2018dynamic,qi2018low}. Thus, the $k$-th output of the classifier for a training example $x_i$ can be defined as
\begin{equation} \label{eq:classifier}
C_k\left(f\left(x_i ; \Theta^{1:B}\right)\right) = \text{softmax} \left( \tau \bar{f}_i^T \overline{w}_k \right),
\end{equation}
where $\bar{f}_i$ is the L2 normalized feature for $x_i$ and $\overline{w}_k$ is the L2 normalized weight for the $k$-th class. $\tau$ is used as a scale parameter for stabilized training \cite{gidaris2018dynamic, chen2018a}. Based on the definition, $C^{Base}$ is denoted as the classifier using base weights, and similarly $C^{Novel}$ is denoted using novel weights.


Then, we consider the mini-batch training with $N_{bs}$ examples and the cost function for our training method is expressed as
\begin{equation} \label{eq:general_loss}
J(\Theta) = \dfrac{1}{N_{bs}} \sum_{i=1}^{N_{bs}} \ell \left( C \left(f\left(\tilde{x}_i ; \Theta^{1:B}\right)\right) ; \tilde{y}_i \right) + R,
\end{equation}
where there exist three components: (a) the virtual training example $\tilde{x}_i$ and label $\tilde{y}_i$, (b) a loss function $\ell$ and (c) a regularizer $R$. We sequentially elaborate on the components in the following subsections.

\subsubsection{Inference}
After training base classes, we report the classification performance on unseen classes that have only a few training examples. We consider that a test dataset has $C^N$ classes, which are disjoint to $C^B$ classes for a training dataset. Thus, this inference process measures how well the network trained on base classes is generalized to unseen classes. For this measurement, we randomly sample $n$ classes from $C^N$ classes, and pick $k$ examples from each class. The typical numbers for few-shot learning are $n=5$ and $k=1$ or $5$. This setting is called $n$-way $k$-shot classification.

After the sampling process, we generate the weight of the $j$-th unseen class as follows:
\begin{equation} \label{Eq: novel_weights}
w^N_j = \dfrac{1}{k} \sum_{i=1}^{k} f_{i,j},
\end{equation}
where $f_{i,j}$ is the feature of the $i$-th example given for the $j$-th unseen class. Then, a query $x_q$ is classified as
\begin{equation}
\argmax_{k} C_k^{Novel} \left(f\left(x_q ; \Theta^{1:B}\right)\right),
\end{equation}
where $C_k^{Novel}$ is defined in Eq. (\ref{eq:classifier}) with the above novel weights.
We iterate these sampling and inference processes several times to obtain the 95\% confidence interval.

\subsection{Self-Augmentation}
To improve the generalization performance, we propose a training framework called self-augmentation, which consolidates self-mix and knowledge distillation. Self-mix randomly picks a region of an input image and substitutes the pixels of the region into other values of same image. We incorporate the self-mix in the few-shot learning problem and show that the generalization performance can be significantly boosted when collaborating with knowledge distillation. The overall architecture of the proposed method is shown in Fig. \ref{fig:Overall_arc}.

\subsubsection{Self-Mix} \label{sect:method_sm}
Self-mix is applied to a raw input image to produce a transformed virtual example as follows: 
\begin{equation*}
\begin{gathered}
\tilde{x} =  T \left(  x_i \right), \\
\end{gathered}
\end{equation*}
where $T: x_i[P_1] \rightarrow x_i[P_2]$ denotes by the abuse of notation,  the patch $P_1$ of $x_i$ is replaced by the patch $P_2$ of $x_i$. To be specific, we firstly sample a cropping region $P_{1} = (r_{a_1},r_{b_1},r_{w},r_{h})$ from an image. The x-y coordinates  $(r_{a_1},r_{b_1})$ is sampled randomly and $(r_{w},r_{h})$ is set to a predefined size. If the patch exceeded the image boundary, we crop it. Then, a patch $P_{2}$ is sampled with the fixed $(r_{w},r_{h})$ and $(r_{a_2},r_{b_2}) \left( \neq (r_{a_1},r_{b_1}) \right)$ randomly chosen by ensuring not exceeding the image boundary.

\subsubsection{Self-Distillation} \label{sect:method_sa}
Knowledge distillation has been studied to mitigate the overfitting problem by regularizing the posterior probability \cite{hinton2015distilling,zhang2018deep,zhu2018knowledge,sun2019deeply}. Although a recent work showed that knowledge distillation among multiple networks can ease off the high-variance characteristic in few-shot learning \cite{dvornik2019diversity}, this method requires $20$ networks for the best performance. Thus, we incorporate self-distillation into our training framework, which employs auxiliary classifiers \cite{zhu2018knowledge,sun2019deeply}. The concept is to create independent predictions for an input image and share the information that has been learned by each classifier. To ensure that the auxiliary classifiers share their own information, we apply the Kullback--Leibler (KL) divergence as a regularizer $R$ \cite{sun2019deeply}. In summary, the general form in Eq. (\ref{eq:general_loss}) can be modified for our training framework as follows:
\begin{align} \label{eq:final_loss}
J(\Theta) &= \nonumber \dfrac{1}{N_{bs}} \sum_{i=1}^{N_{bs}} \sum_{j=1}^{N_{cls}} \ell \left( C_j^{Base} \left(f \left(\tilde{x}_i ; \Theta^{1:l-1} \cup \Theta^{l:B}_j \right)\right) ; \tilde{y}_i \right) \nonumber \\ 
&+ \dfrac{1}{2N_{cls}}\sum_{i=1}^{N_{cls}} \sum_{\substack{j=1, \\ j \neq i}}^{N_{cls}} D_{KL} \left( C_i^{Base} || C_j^{Base} \right).
\end{align}
Here, $N_{cls}$ is the number of auxiliary classifiers, and for mathematical simplicity we regard the main classifier as one of the auxiliary classfiers. We use the cross-entropy loss for $\ell$. $\Theta^{1:l-1} \cup \Theta_j^{l:B}$ means that the parameters before the $l$-th block are shared among the auxiliary classifiers and the $l:B$ blocks are learned independently for the $j$-th classifier.

\subsubsection{Discussion}
Here, we further discuss the effectiveness of the proposed self-mix and the motivation of auxiliary classifiers as follows.

As regional dropout chooses a random region of an input image and replaces the pixels for other values, it perturbs the data statistics. This prevents the network from memorizing the data statistics of base classes and improves the generalization performance for unseen classes. Meanwhile, there exist two present works \cite{yun2019cutmix, devries2017improved} as regional dropout techniques and have its own disadvantages. Cutmix \cite{yun2019cutmix} exchanges two randomly selected patches from two images, thus encouraging the network to learn two labels simultaneously. However, it has been reported that such label smoothing impairs the ability of knowledge distillation \cite{muller2019does}. Considering that our proposed framework employs knowledge distillation, it is less effective for cutmix to exploit the full capacity of our framework. On the other hands, cutout \cite{devries2017improved} converts the pixels of the region into zeros, which inherently leads to information loss. To solve these problems, we propose self-mix, which exchanges the locations of the patches of an input image itself. Given that self-mix does not have any information loss and label smoothing issues, we find that it generates a synergy effect with knowledge distillation.

Next, several works have found that deep networks are prone to over-confident predictions, and this hinders a network from learning generalization \cite{pereyra2017regularizing,zhang2018deep,thulasidasan2019mixup}. In other words, it is possible that an over-confident network results in a decision boundary that is sharp \cite{verma2018manifold} as  highly optimized for the statistics of a training dataset. However, unseen classes are not guaranteed to follow the distribution of training examples for base classes, and a sharp boundary is likely to produce unstable predictions for two slightly different examples of an unseen class. Thus, to alleviate the possible sudden jumps, we employ auxiliary classifiers that share their own information. This helps an optimizer to search for wide valleys \cite{zhang2018deep}. Also, we found that the explicit regularization about the softmax output produces better generalization ability than regional dropout as an input perturbation.

\begin{figure*}[t]
\begin{center}
\includegraphics[width=1.0\linewidth, height=42mm]{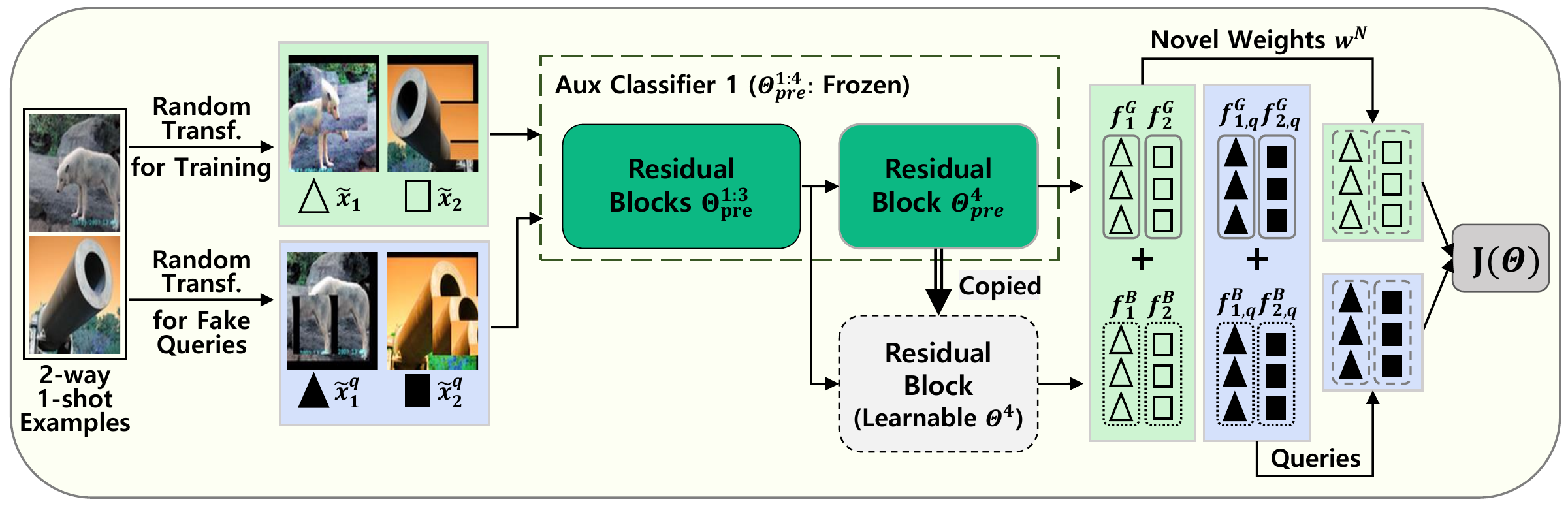}
\end{center}
   \caption{\textbf{Overview of our local representation learning procedure.} We conceptualize our method with an example of $2$-way $1$-shot learning. To fine-tune a network, we randomly apply transformations such as random crop, horizontal flip and regional dropout twice. The first application is intended to produce novel weights $w^N$, and the other is to generate fake queries. These two features yield a loss for a network to be fine-tuned. We copy the last convolutional block of the network and the parameters of the block will be learned to generate a bias $f_i^B$. The role of this bias term is to separate classes that are close to each other. Best viewed in color.}
\label{fig:ft}
\end{figure*}

\subsection{Local Representation Learner}
We have proposed how to train a network on base classes to produce global representations, which can be generalizable to unseen classes. In the test stage, we have $n$-way $k$-shot training examples and $T$ queries for unseen classes. Thus, we now present how to fine-tune the global representations to yield local representations adjusted for the $n \cdot k$ examples. The overall concept is illustrated in Fig. \ref{fig:ft}.

\subsubsection{Preliminary}
For fine-tuning, random transformations are applied on training examples to produce novel weights and fake queries as  follows:
\begin{align*}
\{x_1, x_2, \cdots, x_{n \cdot k}\}  &\xrightarrow[\text{for Training}]{\text{Random Transf.}} \{\tilde{x}_1, \tilde{x}_2, \cdots, \tilde{x}_{n \cdot k}\} \\
\{x_1, x_2, \cdots, x_{n \cdot k}\}  &\xrightarrow[\text{for Fake Queries}]{\text{Random Transf.}} \{\tilde{x}_1^q, \tilde{x}_2^q, \cdots, \tilde{x}_{n \cdot k}^q\},
\end{align*}
where $\tilde{x}_i$ is used to create a novel weight and $\tilde{x}_i^q$ is used to induce a loss. It is worth noting that we only have access to the $n \cdot k$ examples, and we are never informed about the real queries.

\subsubsection{Training}
Our objective is not to destroy the well-learned global representations and we promise to be more discriminative after fine-tuning. Thus, we clone the last block of the pre-trained network and only fine-tune the cloned block. The features extracted from the separate networks are denoted as
\begin{equation*}
\begin{gathered}
f_i^{Global} := f \left( \tilde{x}_i ; \Theta_{pre}^{1:B} \right) \\
f_i^{Bias}   := f \left( \tilde{x}_i ; \Theta_{pre}^{1:B-1} \cup \Theta^B \right),
\end{gathered}
\end{equation*}
where $\Theta_{pre}$ denotes the pre-trained parameters for the base classes. Local representation is defined as the sum of the above two features. Similarly, the features for queries can be defined as $f_{i,q}^{Global}$ and $f_{i,q}^{Bias}$. Then, according to Eq. (\ref{Eq: novel_weights}), the weight for the $j$-th unseen class is produced by
\begin{equation}
w^N_j = \dfrac{1}{k} \sum_{i=1}^{k} \left( f_{i,j}^{Global} + f_{i,j}^{Bias} \right).
\end{equation}
As we have formed novel weights and features for fake queries, a cost function can be defined as
\begin{multline} \label{eq: ft}
J(\Theta) = \dfrac{1}{n \cdot k} \sum_{i=1}^{n \cdot k} \ell \left( C^{Novel} \left( f_{i,q}^{Global}+f_{i,q}^{Bias} \right) ; \tilde{y}_i \right) + \\
\gamma \sum_{j=1}^{n} \sum_{i=1}^{k} \parallel f_{i,j}^{Global} - f_{i,j}^{Bias} \parallel_2,
\end{multline}
where the regularizer $\gamma$ prevents the fine-tuned block $\Theta^B$ from destroying the well-learned feature space given that only a few training examples are available. Overall, we try to learn the bias term to increase the distance between classes that are close to each other so that they are more distinguishable.

\subsubsection{Inference}
A query is classified by the trivial softmax output, but this time we use $T$ real queries. Our proposed fine-tuning method can be applied to any global representations trained on base classes.

\subsection{Experimental Setup}
\subsubsection{Datasets}
\textit{Mini}ImageNet \cite{vinyals2016matching} consists of $100$ classes randomly selected from ILSVRC-2012  \cite{ILSVRC15} and each class has $600$ images, each sized $84 \times 84$. We follow the split proposed in \cite{ravi2016optimization}, namely $64, 16$ and $20$ classes for training, validation and testing, respectively. \textit{Tiered}ImageNet \cite{ren2018meta} has $608$ classes randomly selected from ILSVRC-2012 \cite{ILSVRC15} and these classes are grouped into 34 higher level categories. They are then split into $20, 6$ and $8$ categories to further build $341$, $91$ and $160$ classes for training, validation and testing, respectively. A much larger number of images (totally $779,165$ images) are sized $84 \times 84$. 

\subsubsection{Evaluation}
We report the performance averaged over $2,000$ randomly sampled tasks from the test set to obtain the $95\%$ confidence interval. We use $T=15$ test queries for the $5$-way $5$-shot and the $5$-way $1$-shot, as in \cite{vinyals2016matching,snell2017prototypical,ravi2016optimization}.

\subsubsection{Implementation Details}
For all the datasets, we report the results using ResNet-12 \cite{lee2019meta}, which has four blocks. Each block consists of three $3 \times 3$ Convolution-BatchNorm-LeakyReLU (0.1) and one $2 \times 2$ max pooling. The depths of the four blocks are $64 \rightarrow 160 \rightarrow 320 \rightarrow 640$. 
In \textit{mini}ImageNet, we trained a network for 60 epochs (each epoch consisted of $1,000$ iterations). Initial learning rate was 0.1 and decreased to 0.006, 0.0012 and 0.00024 at 20, 40 and 50 epochs, respectively. In \textit{tiered}ImageNet, the network was trained for 100 epochs (each epoch consisted of $2,000$ iterations). Initial learning rate was 0.1 and decreased to 0.006, 0.0012 and 0.00024 at 40, 80 and 90 epochs, respectively.

\newcommand{\factorial}{\ensuremath{\mbox{\sc Factorial}}}
\begin{algorithm}[!t]
\caption{Pseudo-Code of Self-mix}\label{pcode}
 \hspace*{\algorithmicindent} \textbf{Input} Image with size C$\times$ W$\times$ H  \\
   \hspace*{\algorithmicindent} \textbf{Length} Patch size 
\begin{algorithmic} [1]
\Function{Selfmix}{Input, Length}
    \State H = Input.size(2)
    \State W = Input.size(1)
    \State $x$ = randint(0,W)
    \State $y$ = randint(0,H)
    
    \State $x_1$ = Clip($x$ - Length/2, 0, W)
    \State $x_2$ = Clip($x$ + Length/2, 0, W)
    \State $y_1$ = Clip($y$ - Length/2, 0, H)
    \State $y_2$ = Clip($y$ + Length/2, 0, H)

    \While {true}
    
    \State $x_n$ = randint(0+($x_2-x_1$)/2,W-($x_2-x_1$)/2)
    \State $y_n$ = randint(0+($y_2-y_1$)/2,H-($y_2-y_1$)/2)

    \If{$y_n != y_1$ or $x_n !=x_1$}
        \State break;
    \EndIf

    \EndWhile

   \State Input[:,$x_1:x_2,y_1:y_2$] = Input[:, $x_{n}:x_{n}+(x_2-x_1),y_{n}:y_{n}+(y_2-y_1)$]
\EndFunction
\end{algorithmic}
\end{algorithm}

For self-mix, we randomly sampled a cropping region $P_{1} = (r_{x_1},r_{y_1},r_{w},r_{h})$ from an image. Length of the patch $(r_{w},r_{h})$ was set to $(\frac{W}{2}$, $\frac{H}{2})$. Then, a patch $P_2$ from the same input was sampled with randomly chosen $(r_{x_2}, r_{y_2}) \left( \neq (r_{x_1}, r_{y_1}) \right)$ and the same $(r_{w},r_{h})$. The code-level description is shown in Algorithm \ref{pcode}.

For self-distillation, auxiliary classifiers are branched from the $2$nd and $3$rd blocks of ResNet-12. The two auxiliary classifiers have two and one new ResNet blocks, respectively. All branches were initialized independently, which forces the network into learning different posterior distributions. We use stochastic gradient descent (SGD) with a Nesterov momentum of $0.9$ and a weight decay of $0.0005$. We fix the scale parameter for the classifier to $\tau=20$.

For the local representation learner (LRL), we used the SGD optimizer and the model was trained for 200 epochs per a task. The initial learning rates used for our experiments are shown in Table \ref{tab:ft_lr} and were decreased by a factor of 10 at 80, 120, 160 epochs. To generate fake queries and novel weights, we applied horizontal flip, random crop, color jittering and then we further applied regional dropout such as self-mix.

\begin{table}[t]
  \caption{Initial learning rates and regularization parameters for the local representation learner.}
  \vspace{0.2cm}
  \centering
    \resizebox{\columnwidth}{!}{
    \begin{tabular}{c|cc|cc|cc}
    \hline
    \textbf{Method} & \multicolumn{2}{c|}{\textbf{\textit{mini}ImageNet}} & \multicolumn{2}{c|}{\textbf{\textit{tiered}ImageNet}} & \multicolumn{2}{c}{\textbf{\textit{mini}ImageNet} $\rightarrow$ \textbf{CUB}} \\
          & 1-shot & 5-shot & 1-shot & 5-shot & 1-shot & 5-shot \\
    \hline
    \textbf{LR} $\lambda$  & 1.00E-02 & 1.00E-01 & 1.00E-02 & 1.00E-01 & 1.00E-02 & 1.00E-01 \\
    \textbf{Regularizer} $\gamma$  & 1.00E-01 & 1.00E-01 & 1.00E-01 & 1.00E-01 & 1.00E-01 & 1.00E-01 \\
    \hline
    \end{tabular}%
    }
  \label{tab:ft_lr}%
\end{table}%

\begin{table}[t]
\caption{5-way few-shot classification accuracies on \textit{mini}ImageNet and \textit{tiered}ImageNet with 95\% confidence intervals. All accuracy results are averaged over 2,000 tasks randomly sampled from the test set. LRL denotes the local representation learner.}
\vspace{0.2cm}
  \centering
  \resizebox{\columnwidth}{!}{%
  {\huge
  \renewcommand{\arraystretch}{1.00} 
    \begin{tabular}{p{11.44em}c|cccc}
    \hline
    \multicolumn{1}{c|}{\textbf{Method}} & \multicolumn{1}{c|}{\textbf{Backbone}} & \multicolumn{2}{c|}{\textbf{\textit{mini}ImageNet}} & \multicolumn{2}{c}{\textbf{\textit{tiered}ImageNet}} \\
    \multicolumn{1}{c|}{} &       & 1-shot & \multicolumn{1}{c|}{5-shot} & 1-shot & 5-shot \\
    \hline
    \multicolumn{1}{c|}{LEO\cite{rusu2018meta}} & WRN-28-10 & 61.76 $\pm$ 0.08\% & \multicolumn{1}{c|}{77.59 $\pm$ 0.12\%} &  66.33 $\pm$ 0.05\% &  81.44 $\pm$ 0.09\% \\
    \multicolumn{1}{c|}{MTL\cite{sun2019meta}} & ResNet12 & 61.20 $\pm$ 1.80\% & \multicolumn{1}{c|}{75.50 $\pm$ 0.80\%} & -     & - \\
    \multicolumn{1}{c|}{AM3-TADAM\cite{xing2019adaptive}} & ResNet12 & 65.30 $\pm$ 0.49\% & \multicolumn{1}{c|}{78.10 $\pm$ 0.36\%} & 69.08 $\pm$ 0.47\% & 82.58 $\pm$ 0.31\% \\
    \multicolumn{1}{c|}{MetaOptNet\cite{lee2019meta}} & ResNet12 & 62.64 $\pm$ 0.61\% & \multicolumn{1}{c|}{78.63 $\pm$ 0.46\%} & 65.99 $\pm$ 0.72\% & 81.56 $\pm$ 0.53\% \\
    \multicolumn{1}{c|}{DC\cite{lifchitz2019dense}} & ResNet12 & 62.53 $\pm$ 0.19\% & \multicolumn{1}{c|}{78.95 $\pm$ 0.19\%} & -     & - \\
    \multicolumn{1}{c|}{CAM\cite{hou2019cross}} & ResNet12 & 63.85 $\pm$ 0.48\% & \multicolumn{1}{c|}{79.44 $\pm$ 0.33\%} & 69.89 $\pm$ 0.51\% & 84.23 $\pm$ 0.37\% \\
    \multicolumn{1}{c|}{CC+Rotation\cite{gidaris2019boosting}} & WRN-28-10 & 62.93 $\pm$ 0.45\% & \multicolumn{1}{c|}{79.87 $\pm$ 0.33\%} & 70.53 $\pm$ 0.51\% & 84.98 $\pm$ 0.36\% \\
    \multicolumn{1}{c|}{CTM\cite{li2019finding}} & ResNet18 & 64.12 $\pm$ 0.82\% & \multicolumn{1}{c|}{80.51 $\pm$ 0.13\%} & 68.41 $\pm$ 0.39\% & 84.28 $\pm$ 1.73\% \\
    \multicolumn{1}{c|}{Robust 20-dist++\cite{dvornik2019diversity}} & ResNet18 & 63.73 $\pm$ 0.62\% & \multicolumn{1}{c|}{81.19 $\pm$ 0.43\%} & 70.44 $\pm$ 0.32\% & 85.43 $\pm$ 0.21\% \\
    \hline
    \multicolumn{1}{c|}{Self-Augmentation} & ResNet12 & 65.27 $\pm$ 0.45\% & \multicolumn{1}{c|}{81.84 $\pm$ 0.32\%} &71.26 $\pm$ 0.50\%  & 85.55 $\pm$ 0.34\% \\
    \multicolumn{1}{c|}{Self-Augmentation + LRL} & ResNet12 & \textbf{65.37 $\pm$ 0.45\%} & \multicolumn{1}{c|}{\textbf{82.68 $\pm$ 0.30\%}} & \textbf{71.31 $\pm$ 0.50\%} & \textbf{86.41 $\pm$ 0.33\%}  \\
    \hline
    \end{tabular}%
  }}
\label{table:2benchmarks}
\end{table}%

\subsection{Comparison with the State-of-the-Art Methods}
We compare the proposed method with the state-of-the-art algorithms. As shown in Table \ref{table:2benchmarks}, self-augmentation with LRL clearly outperforms the others by a large margin. It is worth noting that recent techniques \cite{xing2019adaptive, li2019finding} perform well in certain environments such as 1-shot or 5-shot, or on a certain dataset, while the proposed method works decently in all settings. This indicates that it is worthwhile investigating the generalization ability of the standard supervision in relation to few-shot learning. 

\begin{table}[!t]
\caption{5-way few-shot classification accuracies on the domain shift (\textit{mini}ImageNet $\rightarrow$ CUB) with the 95\% confidence intervals. *We re-implemented the official code \cite{chen2018a} to evaluate 1-shot accuracies and 5-shot accuracies were reported from \cite{chen2018a}. `-' denotes that the performance is not provided by the study.}
\vspace{0.2cm}
  \centering
  {\large
  \resizebox{0.6\columnwidth}{!}{
    \begin{tabular}{lcc}
    \hline
    \multicolumn{1}{c|}{\multirow{2}[0]{*}{\textbf{Method}}} & \multicolumn{2}{c}{\textbf{\textit{mini}ImageNet} $\rightarrow$ \textbf{CUB}} \\
    \multicolumn{1}{c|}{} & 1-shot & 5-shot \\
    \hline
    \multicolumn{1}{c|}{RelationNet$^\ast$ \cite{sung2018learning}} & 36.86 $\pm$ 0.70\% & 57.71 $\pm$ 0.73\% \\
    \multicolumn{1}{c|}{ProtoNet$^\ast$ \cite{snell2017prototypical}} & 41.36 $\pm$ 0.70\% & 62.02 $\pm$ 0.70\% \\
    \multicolumn{1}{c|}{Linear Classifier$^\ast$ \cite{chen2018a}} & 44.33 $\pm$ 0.74\% & 65.57 $\pm$ 0.70\% \\
    \multicolumn{1}{c|}{Cosine Classifier$^\ast$ \cite{chen2018a}} & 44.51 $\pm$ 0.80\% & 62.04 $\pm$ 0.76\% \\
    \multicolumn{1}{c|}{Diverse 20 Full \cite{dvornik2019diversity} }& -     & 66.17 $\pm$ 0.55\% \\
    \hline
    \multicolumn{1}{c|}{Self-Augmentation} & 51.50 $\pm$ 0.46\% & 72.00 $\pm$ 0.39\% \\
    \multicolumn{1}{c|}{+ LRL} & \textbf{51.65 $\pm$ 0.46\%} & \textbf{74.20 $\pm$ 0.37\%} \\
    \hline
    \end{tabular}%
}}
 \label{table:cub}
\end{table}%

\begin{figure}[t!]
\centering
\includegraphics[width=0.65\linewidth, height=75mm]{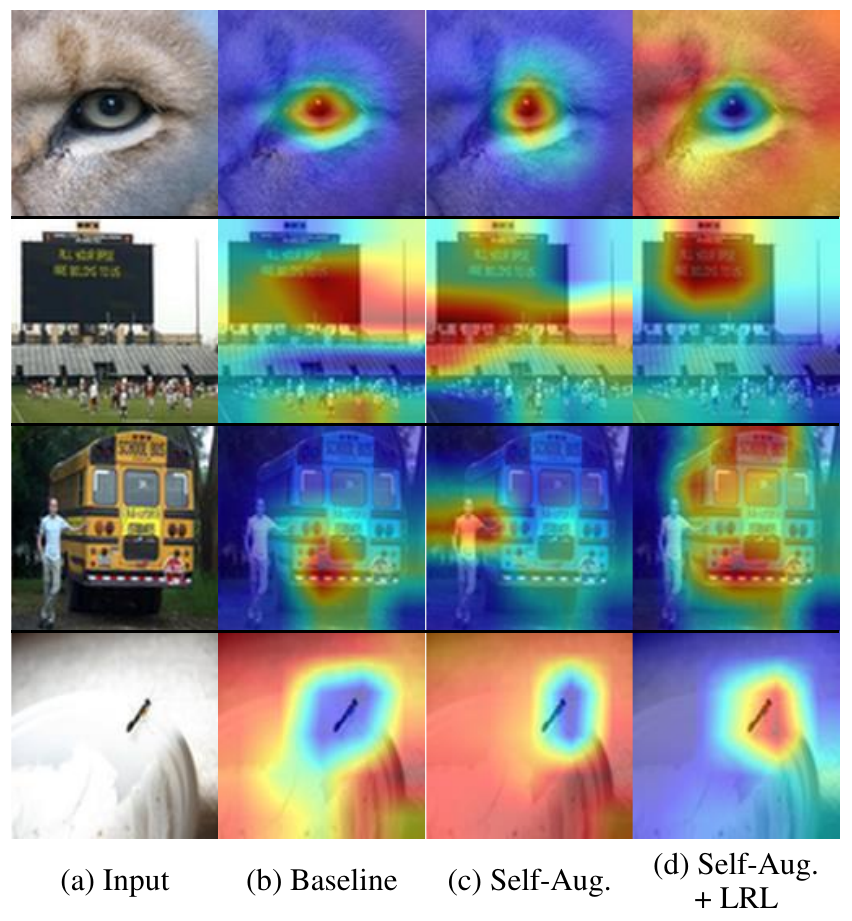}
   \caption{Visualization using the class activation map \cite{zhou2016learning} to show the regions that deep networks focus on.}
\label{fig:vis}
\end{figure} 

\subsection{Domain Shift: \textit{mini}ImageNet to CUB}
We further analyse the generalization ability and the network calibration of the proposed framework. After training a network on \textit{mini}ImageNet, we perform $5$-way classification on CUB \cite{wah2011caltech}. This is a challenging problem as (1) CUB is designed for fine-grained image classification with $200$ bird species, (2) the distributions of the two datasets are largely different and (3) we only have $1$ or $5$ training examples for few-shot learning. With those difficulties, Table \ref{table:cub} shows that self-augmentation significantly surpasses the previous works \cite{sung2018learning,snell2017prototypical,chen2018a,dvornik2019diversity}.

\begin{table*}[t]
\caption{Ablation study on \textit{mini}ImageNet, \textit{tiered}ImageNet and cross-domain benchmarks. Baseline refers to a vanilla network without any regional dropout techniques. SD and SA denotes self-distillation and self-augmentation, respectively.}
\vspace{0.2cm}
\resizebox{\linewidth}{!}{
{\Huge
  \centering
    \begin{tabular}{c|cc|cc|cc}
    \hline
    \textbf{Method} & \multicolumn{2}{c|}{\textbf{\textit{mini}ImageNet}} & \multicolumn{2}{c|}{\textbf{\textit{tiered}ImageNet}} & \multicolumn{2}{c}{\textbf{\textit{mini}ImageNet} $\rightarrow$ \textbf{CUB}} \\
          & 1-shot & 5-shot & 1-shot & 5-shot & 1-shot & 5-shot \\
    \hline
    Baseline  & 61.42 $\pm$ 0.45\%  & 78.32 $\pm$ 0.33\% & 68.22 $\pm$ 0.50\%  & 83.21 $\pm$ 0.36\% &  47.76 $\pm$ 0.44\%  & 67.40 $\pm$ 0.38\% \\
    \hline
    \rowcolor{Gray}
    Cutout & 62.38 $\pm$ 0.44\%  & 79.18 $\pm$ 0.33\% & 69.40 $\pm$ 0.51\%  & 84.27 $\pm$ 0.36\% &  47.46 $\pm$ 0.44\%  & 67.79 $\pm$ 0.40\% \\
    Cutmix & 62.81 $\pm$ 0.45\%  & 79.82 $\pm$ 0.33\% & 69.09 $\pm$ 0.49\%  & 84.21 $\pm$ 0.35\% &  48.35 $\pm$ 0.44\%  & 67.77 $\pm$ 0.39\% \\
    \rowcolor{Gray}
    Selfmix & 62.85 $\pm$ 0.45\%  & 79.83 $\pm$ 0.32\% & 69.95 $\pm$ 0.40\%  & 84.39 $\pm$ 0.35\% & 48.73 $\pm$ 0.45\% & 69.20 $\pm$ 0.39\% \\
    \hline
    Self-Distillation & 63.11 $\pm$ 0.45\% & 79.93 $\pm$ 0.33\% & 70.05 $\pm$ 0.49\% & 84.92 $\pm$ 0.34\% & 48.91 $\pm$ 0.44\% & 69.45 $\pm$ 0.38\% \\
    \rowcolor{Gray}
    SD + Cutout & 64.61 $\pm$ 0.44\% & 81.57 $\pm$ 0.31\% & 70.76 $\pm$ 0.50\% & 85.50 $\pm$ 0.35\% & 48.94 $\pm$ 0.43\% & 69.65 $\pm$ 0.39\% \\
    \rowcolor{Gray}
    SD + Cutout + LRL & 64.93 $\pm$ 0.45\% & 82.34 $\pm$ 0.30\% & 70.82 $\pm$ 0.50\% & 86.15 $\pm$ 0.33\% &   48.93 $\pm$ 0.42\%  & 73.37 $\pm$ 0.36\% \\
    SD + Cutmix & 64.44 $\pm$ 0.45\% & 81.58 $\pm$ 0.32\% & 70.46 $\pm$ 0.49\% & 85.51 $\pm$ 0.34\% & 50.43 $\pm$ 0.45\% & 70.70 $\pm$ 0.39\% \\
    SD + Cutmix + LRL & 64.67 $\pm$ 0.45\% & 81.52 $\pm$ 0.31\% & 70.48 $\pm$ 0.48\% & 85.60 $\pm$ 0.34\% & 49.88 $\pm$ 0.43\% & 72.35 $\pm$ 0.37\% \\
    \hline
    \rowcolor{Gray}
    \textbf{Self-Augmentation} & \textbf{65.27 $\pm$ 0.45\%} & \textbf{81.84 $\pm$ 0.32\%} & \textbf{71.26 $\pm$ 0.50\%} & \textbf{85.55 $\pm$ 0.34\%} & \textbf{51.50 $\pm$ 0.46\%} & \textbf{72.00 $\pm$ 0.39\%} \\
    \rowcolor{Gray}
    \textbf{SA + LRL} & \textbf{65.37 $\pm$ 0.45\%} & \textbf{82.68 $\pm$ 0.30\%} & \textbf{71.31 $\pm$ 0.50\%} & \textbf{86.41 $\pm$ 0.33\%} & \textbf{51.65 $\pm$ 0.46\%} & \textbf{74.20 $\pm$ 0.37\%} \\
    \hline
    \end{tabular}}}%
\label{table:self-aug}
\end{table*}%

\subsection{Ablation Study} \label{sect:perf_sa}
\subsubsection{Effect of the Local Representation Learner}
Fig. \ref{fig:vis} shows that there exist cases where the local representation learner (LRL) fixes the deep network to focus on more discriminative parts. As a result, only self-augmentation with LRL correctly classifies the below images. This indicates that a network can be further enhanced even with a few training examples using a carefully designed strategy.

\subsubsection{Comparison with Various Regional Dropout Techniques}
As we proposed the framework consolidating regional dropout and self-distillation, Table \ref{table:self-aug} shows that how performance changes by adopting various regional dropout methods and self-distillation. Baseline refers to a network using light augmentation such as random color jittering, cropping and horizontal flipping. The results indicate four notable aspects: (1) Self-augmentation significantly outperforms the baseline using light augmentation only. (2) Although either regional dropout or self-distillation can improve the generalization capability, exploiting both methods leads to higher performance gains. (3) As discussed in Sect. \ref{sect:method_sm}, the proposed self-mix has a synergistic effect with self-distillation as it does not require pixel removal \cite{devries2017improved} or mixed labels \cite{yun2019cutmix}. (4) When using cutmix \cite{yun2019cutmix} for the local representation learner, the performance remains almost the same. As only a few training examples exist, we conjecture that the mixed labels produced by cutmix increase the complexity of fine-tuning. To sum up, although several regional dropout techniques have been studied, self-mix is more flexible to be used with distillation or local representation leaning.

\begin{table*}[t]
    \caption{Effect of label smoothing on \textit{mini}ImageNet. Applying label smoothing to each method decreases their original performances for unseen classes. }
    \vspace{0.2cm}
\resizebox{\columnwidth}{!}{
  \centering
    \begin{tabular}{c|cc|cccccc}
    \hline
    \textbf{Method} & \multicolumn{2}{c|}{\textbf{Baseline}} & \multicolumn{2}{c|}{\textbf{Baseline}} & \multicolumn{2}{c|}{\textbf{Self-Distillation}} & \multicolumn{2}{c}{\textbf{Self-Augmentation}} \\
    \multicolumn{1}{c|}{} & \multicolumn{2}{c|}{} & 1-shot & \multicolumn{1}{c|}{5-shot} & 1-shot & \multicolumn{1}{c|}{5-shot} & 1-shot & 5-shot \\
    \hline
    \textbf{Test class} & \multicolumn{2}{c|}{Base class} & \multicolumn{6}{c}{Unseen class} \\
    \hline
    \textbf{w/o label smoothing} & \multicolumn{2}{c|}{80.22\%} & 61.42\% & 78.32\% & 63.11\% & 79.93\% & 65.27\% & 81.84\% \\
    \textbf{w label smoothing} & \multicolumn{2}{c|}{81.36\%} & 61.27\% & 77.03\% & 61.96\% & 77.45\% & 63.29\% & 78.24\% \\
    \hline
    \textbf{Gain} & \multicolumn{2}{c|}{\textcolor[rgb]{ 0,  .502,  .502}{(+1.14\%)}} & \textcolor[rgb]{ 1,  0,  0}{(-0.15\%)} & \textcolor[rgb]{ 1,  0,  0}{(-1.29\%)} & \textcolor[rgb]{ 1,  0,  0}{(-1.15\%)} & \textcolor[rgb]{ 1,  0,  0}{(-2.48\%)} & \textcolor[rgb]{ 1,  0,  0}{(-1.84\%)} & \textcolor[rgb]{ 1,  0,  0}{(-3.60\%)} \\
    \hline
    \end{tabular}%
    }

  \label{table:ablation}%
\end{table*}%

\begin{figure*}[t]
\begin{center}
\includegraphics[width=1.0\linewidth, height=50.5mm]{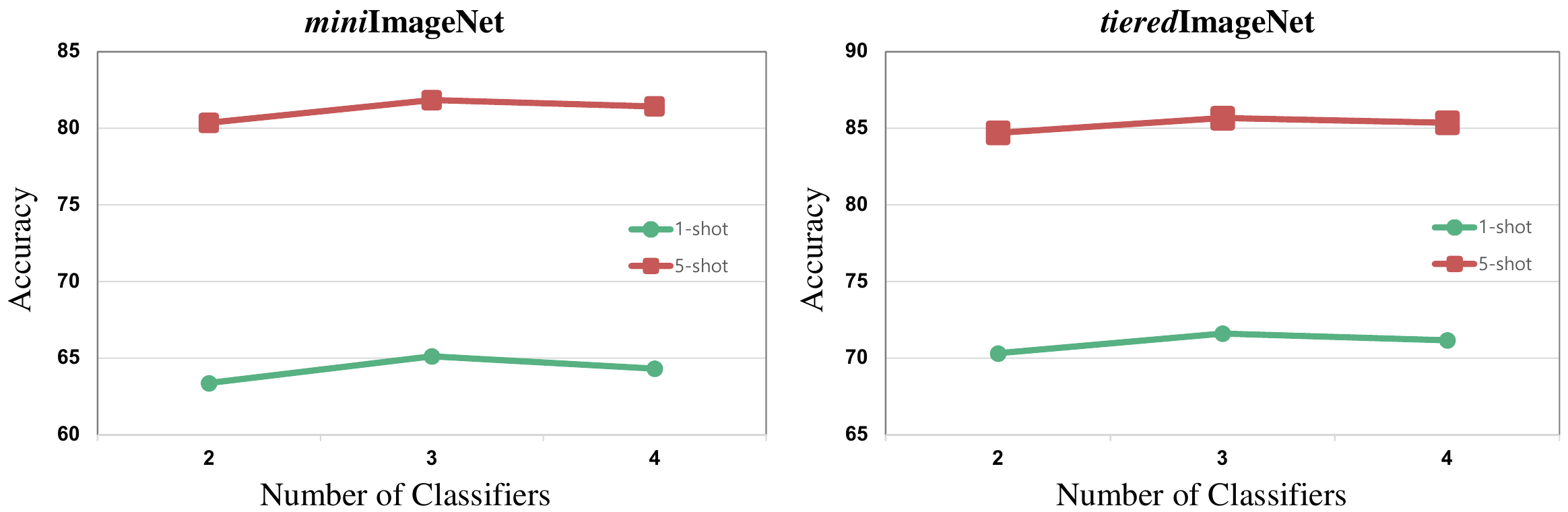}
\end{center}
   \caption{Test accuracies (\%) with various numbers of classifiers for self-distillation. In both cases, using three classifiers shows the highest accuracy.}
\label{fig:ncls}
\end{figure*}

\subsubsection{Effect of Label Smoothing}
As we deal with a memorization problem of deep networks in terms of few-shot learning, we further present the performance with label smoothing that is another way to perturb output distributions. Though it is well-known that label smoothing is beneficial for standard classification problems \cite{szegedy2015going}, Table \ref{table:ablation} indicates that label smoothing is not effective for few-shot learning and there exist a significant performance drop with self-distillation. Furthermore, it is worth noting that Table \ref{table:self-aug} also shows that cutmix, which learns two labels simultaneously similar to label smoothing, has less performance gain when using self-distillation.

\subsubsection{Number of Classifiers}
Fig. \ref{fig:ncls} shows that how different numbers of classifiers $N_{cls}$ in Eq. \ref{eq:final_loss} affect the classification performance. We can verify that there exists an optimal number of classifiers and this can be seen as the trade-off between the amount of knowledge sharing and the complexity of the parameter space.

\section{Conclusion}
In this paper, we show that unseen classes with a few training examples can be classified with a standard supervised training. Especially, we aim at generalizing deep networks to unseen classes by alleviating the memorization phenomenon, which is less studied for few-shot learning. To achieve this, we design a framework using regional dropout and self-distillation to perturb the input and output information. Especially, we show that the newly proposed regional dropout, called self-mix, produces state-of-the-art results when cooperating with self-distillation. We also present a local representation learner to exploit a few training examples of unseen classes, which improves the performance for all few-shot learning benchmarks and especially works well on a cross-domain task. More importantly, we show that existing perturbation methods, which are designed for a standard classification setting, such as cutmix, cutout and label smoothing are not the optimal choices for few-shot learning as they are not flexible enough to be used with knowledge distillation or local representation leaning.

\section*{Acknowledgment}
This work was supported by Institute for Information \& communications Technology Planning \& Evaluation(IITP) grant funded by the Korea government(MSIT) (No. 2017-0-01779, A machine learning and statistical inference framework for explainable artificial intelligence, No. 2019-0-01371, Development of brain-inspired AI with human-like intelligence, and No. 2019-0-00079, Department of Artificial Intelligence(Korea University)).

\bibliography{Self_Augmentation_Fewshot_Learning}

\end{document}